\documentclass[letterpaper,10pt,conference]{ieeeconf}
\IEEEoverridecommandlockouts
\usepackage{cite}
\usepackage{amsmath,amssymb,amsfonts}
\usepackage{graphicx}
\usepackage{stfloats}
\usepackage{balance}
\usepackage{textcomp}
\usepackage{xcolor}
\makeatletter
\let\NAT@parse\undefined
\makeatother
\usepackage{hyperref}
\ifdefined\pdfminorversion\pdfminorversion=6\fi
\title{\LARGE \bf Polarization calibration of fisheye cameras using skylight in the wild}

\author{Shuo Wang, Shiqi Che, Dianze Dong, and Wenxin Liu$^{*}$%
\thanks{This work was supported by the National Natural Science Foundation of China under Grant T2596034.}%
\thanks{Shuo Wang, Shiqi Che, Dianze Dong, and Wenxin Liu are with the School of
Automation Science and Electrical Engineering, Beihang University, Beijing
100191, China (E-mail: sWangsuki@buaa.edu.cn; sqche@buaa.edu.cn; dzdong@buaa.edu.cn; wenxinl@buaa.edu.cn). }%
\thanks{$^{*}$Wenxin Liu is the corresponding author of this work.}%
}

\begin{document}
\maketitle
\thispagestyle{empty}
\pagestyle{empty}

\begin{abstract}
Skylight polarization provides directional cues for robotic navigation, but fisheye lens effects and analyzer imperfections can degrade polarization measurements across the field of view. We present a joint lens--analyzer calibration method that explicitly models incidence-dependent lens transmission together with analyzer gains and angular offsets in a unified eight-parameter formulation. Given known geometric calibration, rotations between camera frames can be estimated using visual features and cross-frame directional polarized light consistency can be used for polarization calibration. Therefore, our calibration approach requires only a short handheld outdoor sequence under natural skylight, without dedicated polarization calibration equipment or a predeﬁned sky polarization model. On real-world outdoor sequences, the proposed method achieves a weighted cross-frame angle-of-linear-polarization (AoLP) mean absolute error below 0.5° across all incidence angles, with a 73.4\% reduction in the 80°--90° incidence angle region, and generalizes to an independent sequence with an overall mean absolute error of 0.466° without refitting. Preliminary solar-azimuth experiments using a camera mounted on a quadruped robot further show improved alignment with the visible solar center, supporting the potential of calibrated AoLP measurements for relative-heading estimation and future polarization--inertial navigation. The calibration software tool will be released upon acceptance of this paper.
\end{abstract}

\section{Introduction}
\label{sec:introduction}
Skylight polarization provides a natural compass for robotic navigation by encoding heading relative to the solar meridian. Atmospheric scattering creates a polarization pattern tied to the sun's position, which insects such as bees exploit through specialized receptors in their compound eyes~\cite{gkanias2023celestial}. This directional cue has recently been incorporated into visual-inertial localization frameworks to constrain heading drift~\cite{wan2024visual} and shown significant improvement in GPS-denied conditions. Beyond navigation, polarization sensing has shown promise in transparent object reconstruction~\cite{shao2023transparent}, further expanding its role in robotic perception.

Current polarization navigation systems often employ point-source polarization sensor or compound-eye-inspired configurations that sample a small set of viewing directions~\cite{xue2023self}. Clouds and weak polarization can easily compromise the available directional cues~\cite{xia2023exploration}, while sparse sampling restricts the ability to recover reliable information from other sky regions. On the other hand, a key limitation of image-based polarization navigation is the lack of effective full-field calibration for fisheye polarization cameras. Although fisheye optics provide broad angular coverage of skylight polarization~\cite{ma2015evaluation}, conventional DoFP calibration methods primarily correct analyzer response errors~\cite{powell2013calibration}, leaving substantial polarization distortions in peripheral image regions due to incidence-dependent lens responses~\cite{lane2021calibration,bai2023angle}. Existing image-based navigation approaches typically assume a horizontally mounted, upward-facing camera and estimate heading from zenith-centered polarization patterns or manually selected image regions~\cite{pan2024image}. This reliance on observations near the optical axis limits the use of peripheral information and makes heading estimation sensitive to camera orientation changes caused by platform motion, potentially reducing navigation accuracy.

\begin{figure}[!t]
\centering
\includegraphics[width=\columnwidth]{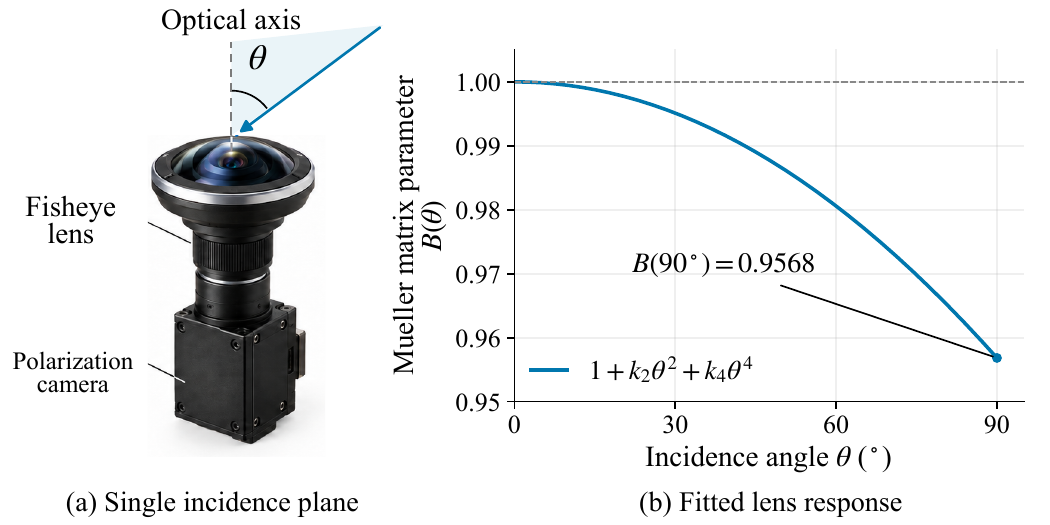}
\caption{Axisymmetric lens response modeled in a single plane of incidence. (a) Fisheye camera geometry and incidence angle $\theta$. (b) Blue-channel Mueller matrix parameter $B(\theta)$.}
\label{fig:blue_fitted_response}
\end{figure}

We propose a joint lens--analyzer calibration framework that combines the incidence-dependent relative transmission of orthogonal polarization components with analyzer gains and angular offsets in a unified forward model. Under the axial-symmetry assumption, we model the lens response in a single incidence plane and extend it across viewing azimuths by rotating the local polarization basis (Fig.~\ref{fig:blue_fitted_response}). The resulting calibration effectively eliminates lens-induced polarization distortions and corrects analyzer imperfections, enabling accurate polarization measurements across the entire image, including peripheral regions.

In addition to the newly proposed model for full-field polarization calibration, we combine the advantages of image-based polarization sensing with visual motion estimation to develop a practical, low-cost calibration procedure. Our method requires only a short handheld outdoor video sequence capturing the sky and surrounding structures near the horizon, without any expensive dedicated laboratory equipment or complicated experimental setup.

\begin{figure*}[!t]
\centering
\includegraphics[width=\textwidth,pagebox=cropbox]{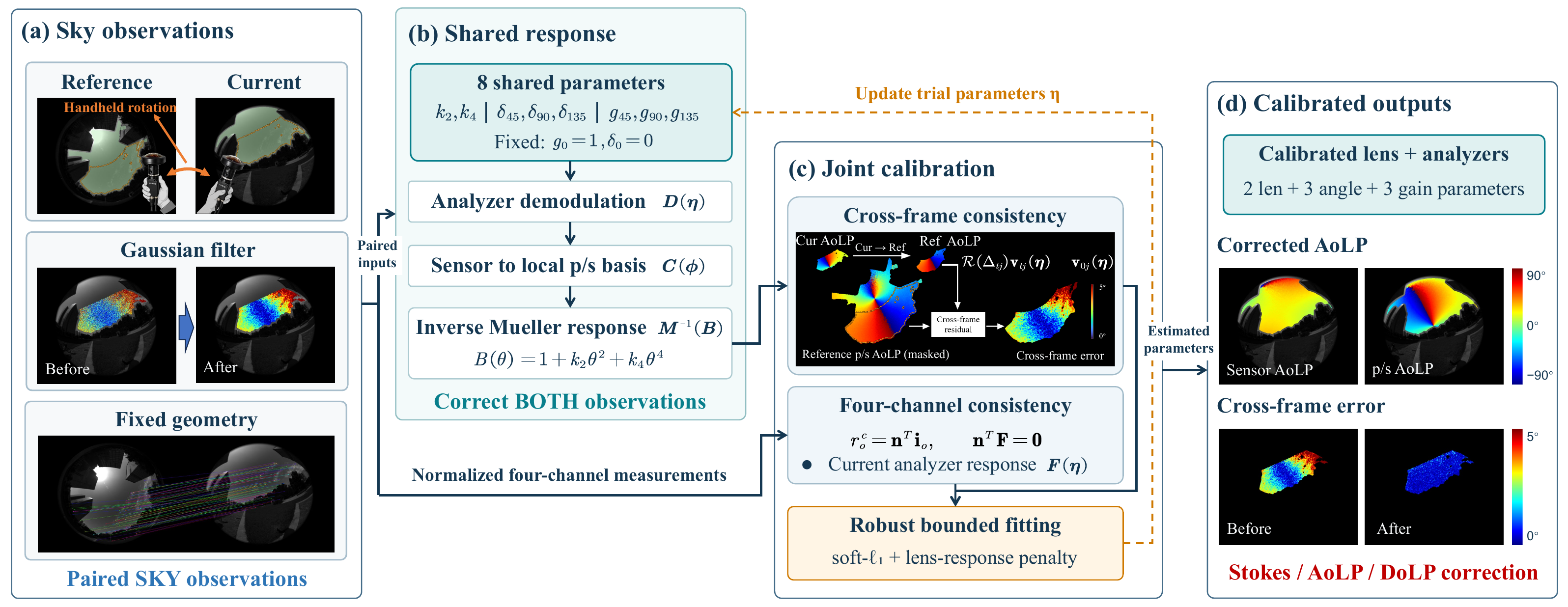}
\caption{Overview of the proposed sky-based joint calibration framework. (a) Sky observations are filtered and paired using fixed geometric correspondences. (b) A shared eight-parameter lens--analyzer model corrects both observations. (c) Robust joint optimization combines cross-frame polarization consistency, four-channel consistency, and a lens-response penalty. (d) The estimated lens and analyzer parameters enable subsequent Stokes, AoLP, and DoLP correction.}
\label{fig:overview}
\end{figure*}

The main contributions of this work are:
\begin{itemize}
\item A unified eight-parameter lens--analyzer response model for fisheye polarization cameras, enabling joint calibration of lens transmission, analyzer response, and angular misalignment.
\item A practical handheld outdoor calibration framework requiring neither dedicated polarization calibration equipment nor predefined sky polarization models.
\item An open-source software package, to be released upon acceptance, that enables convenient use of the proposed calibration system and supports practical deployment. 
\end{itemize}

\section{Related Work}
\label{sec:related_work}

\subsection{DoFP Polarization Camera Calibration}

DoFP calibration typically characterizes sensor nonuniformity through scalar or matrix-based corrections for channel transmission and analyzer orientation errors~\cite{powell2013calibration,wang2024polarization}. Related studies quantify the sensitivity of polarization measurements to component inaccuracies~\cite{lian2025quantitative} and exploit measurement redundancy for analyzer-angle self-calibration~\cite{xue2023self}. These approaches primarily focus on sensor-level calibration and generally do not explicitly model lens-induced polarization distortions. This omission becomes critical for fisheye systems, where incidence-dependent lens transmission varies across the field of view and cannot be corrected by analyzer calibration alone.

\subsection{Wide-field and Natural-Light Polarimetric Calibration}

Wide-field polarimetric calibration requires modeling both geometric distortions and polarization-dependent optical responses. Existing approaches address polarization reference-frame variations caused by omnidirectional projections~\cite{bai2023angle,blocker2024rotation}, while pixel-wise camera response and Mueller matrix formulations characterize the system-level optical effects at specific image locations~\cite{rodriguez2022practical,wan2021measurement}. However, these methods provide limited coverage of wide-field imaging and lack explicit disentanglement of angular lens responses from analyzer imperfections. Recent wide-field camera calibration further highlights this limitation, as theoretical models often neglect lens polarization effects~\cite{weber2024polarization}. Although controlled polarized illumination and specialized setups can improve calibration accuracy~\cite{sun2025wide}, they require dedicated hardware and remain difficult to deploy in the field. Natural-light calibration provides a flexible alternative, but existing approaches have mainly focused on spaceborne scenarios~\cite{avitan2023using}. To the best of our knowledge, a practical and effective calibration method that jointly accounts for lens and analyzer effects across the full field of view of a fisheye polarization camera remains lacking.

\section{Method}

\subsection{Overview}
\label{sec:overall}
Fig.~\ref{fig:overview} summarizes the proposed calibration pipeline. We jointly calibrate the effective lens and analyzer responses of a fisheye DoFP camera from a short rotating sky sequence. Repeated observations of each sky sample share a polarization direction without requiring known Stokes vectors. A calibrated projection and image-estimated rotations establish correspondences. Finally, a shared eight-parameter model is fitted using cross-frame direction consistency and four-channel redundancy.

\subsection{Sky Observation Alignment and Preprocessing}
\label{sec:preprocess}
We exploit the quasi-static property of sky polarization patterns. Due to the large distance of sky regions, translational motion is negligible, allowing the camera motion to be modeled as pure rotation. The relative camera rotations are estimated from SIFT correspondences extracted from total-intensity images using a ray-based RANSAC scheme, and are subsequently fixed during calibration.

Given a reference sky observation $j$ in frame $0$ and its corresponding observation in frame $t$, the geometric correspondence is established as
\begin{equation}
\mathbf r_{0j}=\mathcal P^{-1}(\mathbf x_{0j}),\quad
\mathbf r_{tj}=\mathbf R_{t0}\mathbf r_{0j},\quad
\mathbf x_{tj}=\mathcal P(\mathbf r_{tj}),
\label{eq:correspondence}
\end{equation}
where $\mathcal P$ denotes the calibrated fisheye projection model, $\mathbf x$ represents an image coordinate, $\mathbf r$ is the corresponding unit vector, and $\mathbf R_{t0}\in\mathrm{SO}(3)$ denotes the rotation from the reference camera frame to frame $t$.

To reduce noise and sensitivity to the spatial offsets between analyzer pixels, we apply Gaussian filtering independently to the four raw channels of one color channel under a common valid-sky mask $m_t$:
\begin{equation}
\bar I_{t,k}=\frac{G_\sigma*(m_t I_{t,k})}{G_\sigma*m_t},
\label{eq:filter}
\end{equation}
where $k\in\{0,45,90,135\}$ labels the analyzer channels, $*$ denotes convolution, and $G_\sigma$ is a Gaussian kernel. 

\begin{figure}[!htbp]
\centering
\includegraphics[width=\columnwidth]{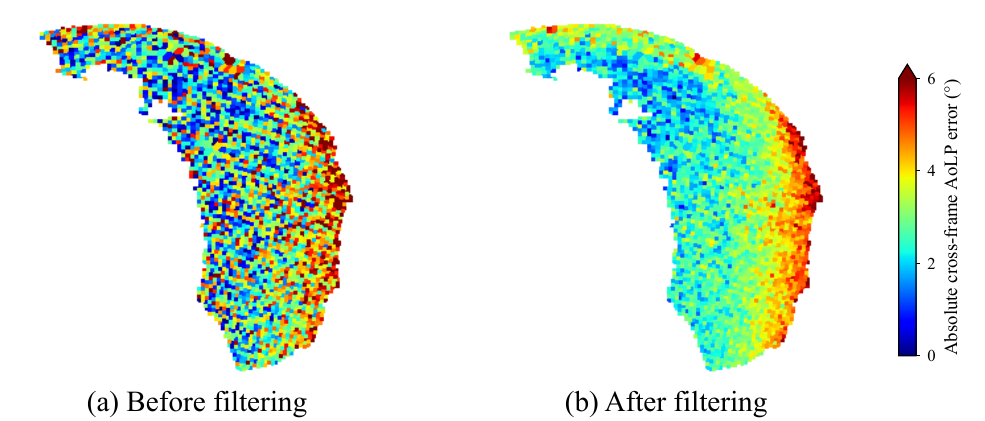}
\caption{Effect of Gaussian filtering on cross-frame AoLP consistency. (a), (d) Absolute cross-frame AoLP errors before and after filtering, respectively.}
\label{fig:filtering_consistency}
\end{figure}

For linear Stokes $\mathbf S=[S_0,S_1,S_2]^T$, where $S_0=\tfrac12(\bar I_{t,0}+\bar I_{t,45}+\bar I_{t,90}+\bar I_{t,135})$ is total intensity, $S_1=\bar I_{t,0}-\bar I_{t,90}$ and $S_2=\bar I_{t,45}-\bar I_{t,135}$ are the linear-polarization components. DoLP and AoLP are $d=\sqrt{S_1^2+S_2^2}/S_0$ and $\alpha=\tfrac12\operatorname{atan2}(S_2,S_1)$ in the stated basis. Mask-normalized filtering removes invalid neighbors and reduces AoLP fluctuations by suppressing intensity inconsistencies prior to Stokes reconstruction as shown in Fig.~\ref{fig:filtering_consistency}.

\subsection{Lens--Analyzer Joint Response Model}
\label{sec:response}

To account for both lens-induced polarization distortion and analyzer imperfections, we introduce a joint lens--analyzer response model that incorporates the optical and sensor responses within a unified formulation.

\textbf{Analyzer response.}
The response of four-channel polarization analyzer is represented by $\mathbf F\in\mathbb R^{4\times3}$, which maps the Stokes vector to measured channel intensities. 
For the $k$-th analyzer channel, the corresponding row of $\mathbf F$ is defined as
\begin{equation}
\mathbf F_{k,:}=\frac{g_k}{2}
[1,\cos2\beta_k,\sin2\beta_k],\quad
\beta_k=\beta_k^{\mathrm{nom}}+\delta_k,
\label{eq:analyzer}
\end{equation}
where $\beta_k^{\mathrm{nom}}\in\{0,\pi/4,\pi/2,3\pi/4\}$ denotes the analyzer orientation. The parameters $g_k$ and $\delta_k$ represent the gain correction and angular offset of the $k$-th analyzer channel, respectively. Since only relative channel gains and analyzer angular offsets are identifiable, the $0^\circ$ analyzer is selected as the reference by fixing $g_0=1$ and $\delta_0=0$.

\textbf{Lens response.}
The fisheye lens introduces polarization-dependent transmission, resulting in view-dependent distortions of the measured polarization state. 
To account for this effect, we parameterize the lens response as a function of the incidence angle and jointly optimize it with the analyzer response during calibration.

For a unit ray $\mathbf r=[r_x,r_y,r_z]^T$, the incidence and azimuth angles are defined as
$\theta=\arccos(r_z)$ and $\phi=\operatorname{atan2}(r_y,r_x)$, respectively. Given the optical axis $\mathbf z=[0,0,1]^T$, we construct a local polarization basis
$\mathbf E(\mathbf r)=[\mathbf p,\mathbf s]$, where
$\mathbf p=(\mathbf z-r_z\mathbf r)/\|\mathbf z-r_z\mathbf r\|$
denotes the component parallel to the incidence plane and $\mathbf s=\mathbf r\times\mathbf p$ is the orthogonal component.

Assuming an axisymmetric and non-depolarizing lens with negligible relative phase delay, the lens-induced polarization transformation in this basis is modeled by the Jones matrix
\begin{equation}
\mathbf J(\theta)=\operatorname{diag}(1,B(\theta)),\qquad
B(\theta)=1+k_2\theta^2+k_4\theta^4,
\label{eq:B}
\end{equation}
where $B(\theta)$ is the relative $s$-to-$p$ amplitude transmission within the local incidence plane and parameterizes the lens Mueller matrix below. The even-order polynomial provides a compact angular-dependent model while satisfying $B(0)=1$, which ensures equal transmission along the optical axis. The lens parameters $k_2$ and $k_4$ are jointly estimated with the analyzer gains and angular offsets during calibration.

To account for partially polarized skylight, the lens response is represented in the linear-Stokes domain using the Mueller matrix
\begin{equation}
\mathbf M(B)=\frac12
\begin{bmatrix}
1+B^2 & 1-B^2 & 0\\
1-B^2 & 1+B^2 & 0\\
0 & 0 & 2B
\end{bmatrix},
\label{eq:mueller}
\end{equation}
where $B=B(\theta)$. 
This Mueller representation extends the Jones-based model to partially polarized light by jointly modeling variations in the degree of polarization and the angle of linear polarization.

\textbf{Forward model and correction.}
For observation $o$, the normalized four-channel intensity measurement is defined as $\mathbf i_o=\bar{\mathbf I}_o/S_{0,\mathrm{nom},o}$, where 
$\bar{\mathbf I}_o=[\bar I_{0,o},\bar I_{45,o},\bar I_{90,o},\bar I_{135,o}]^T$
and 
$S_{0,\mathrm{nom},o}=\frac12\sum_k\bar I_{k,o}$ 
is used to remove the common intensity scale.

The transformation between the sensor coordinate basis and the local $p/s$ basis is given by
\begin{equation}
\mathbf C(\phi)=\begin{bmatrix}
1&0&0\\0&\cos2\phi&-\sin2\phi\\0&-\sin2\phi&-\cos2\phi
\end{bmatrix},\qquad \mathbf S_{ps}=\mathbf C(\phi)\mathbf S_s.
\label{eq:coordinate}
\end{equation}
Combining the analyzer response and lens-induced polarization transformation, the measurement model is formulated as
\begin{equation}
\mathbf i_o=\mathbf F\mathbf C(\phi_o)^{-1}
\mathbf M(B(\theta_o))\mathbf S_{\mathrm{in},o}+\boldsymbol\epsilon_o,
\label{eq:full_forward}
\end{equation}
where $\boldsymbol\epsilon_o$ denotes measurement noise. The calibration parameters are $\boldsymbol\eta=[k_2,k_4,\delta_{45},\delta_{90},\delta_{135},g_{45},g_{90},g_{135}]^T\in\mathbb R^8$.

To avoid propagating analyzer calibration errors into the reconstructed Stokes vectors, we incorporate the current analyzer parameters into the Stokes demodulation process. Specifically, the sensor-frame Stokes vector is estimated by minimizing $\|\mathbf i_o-\mathbf F\mathbf S_s\|_2^2$.

Assuming that $\mathbf F$ has full column rank, the closed-form solution is
\begin{equation}
\begin{aligned}
&\mathbf D(\boldsymbol\eta)=(\mathbf F^T\mathbf F)^{-1}\mathbf F^T,\\
&\widehat{\mathbf S}_{s,o}(\boldsymbol\eta)=\mathbf D(\boldsymbol\eta)\mathbf i_o,\\
&\widehat{\mathbf S}_{\mathrm{in},o}(\boldsymbol\eta)
=\mathbf M(B(\theta_o))^{-1}
\mathbf C(\phi_o)\widehat{\mathbf S}_{s,o}(\boldsymbol\eta),
\end{aligned}
\label{eq:correction}
\end{equation}
where $\mathbf D(\boldsymbol\eta)$ is the parameter-dependent demodulation matrix. 
$\widehat{\mathbf S}_{s,o}$ and $\widehat{\mathbf S}_{\mathrm{in},o}$ denote the estimated Stokes vectors in the sensor and incident coordinate systems, respectively.

\subsection{Cross-Frame Polarization Consistency}
\label{sec:crossframe}
The corrected Stokes vectors obtained from~\eqref{eq:correction} are represented in local $p/s$ polarization bases, which vary with both viewing direction and camera pose. 
Consequently, polarization measurements from different frames cannot be directly compared without compensating for the basis change. 
For each cross-frame correspondence, we first transform the polarization representation of the current observation into the local basis of the reference observation.

For a correspondence $(t,j)$ between frame $t$ and the reference frame $0$, the relative transformation between the two local polarization bases is computed as

\begin{equation}
\begin{aligned}
&\mathbf T_{tj}=\mathbf E(\mathbf r_{0j})^{T}
\mathbf R_{t0}^{T}\mathbf E(\mathbf r_{tj}),\\
&\Delta_{tj}=\operatorname{atan2}((\mathbf T_{tj})_{21},
(\mathbf T_{tj})_{11}),
\end{aligned}
\label{eq:transport}
\end{equation}
where $\mathbf T_{tj}\in\mathbb R^{2\times2}$ denotes the in-plane rotation between the two polarization bases and $\Delta_{tj}$ is the corresponding basis rotation angle.

For observation $o$, we represent the corrected linear polarization state using the normalized Stokes direction
\begin{equation}
\mathbf v_o(\boldsymbol\eta)=
\frac{[\widehat S_{1,\mathrm{in},o},\widehat S_{2,\mathrm{in},o}]^{T}}
{\sqrt{\widehat S_{1,\mathrm{in},o}^{\,2}+\widehat S_{2,\mathrm{in},o}^{\,2}}},
\label{eq:direction}
\end{equation}
where $\widehat S_{1,\mathrm{in},o}$ and $\widehat S_{2,\mathrm{in},o}$ are obtained from the corrected incident Stokes vector in~\eqref{eq:correction}. The cross-frame polarization consistency residual is defined as
\begin{equation}
\mathbf r^a_{tj}=\sqrt{w_{tj}}
\left[\mathcal R(\Delta_{tj})\mathbf v_{tj}(\boldsymbol\eta)
-\mathbf v_{0j}(\boldsymbol\eta)\right],
\label{eq:angular_residual}
\end{equation}
where $\mathcal R(\Delta_{tj})$ transports the current observation into the reference polarization basis. 
For two measurements corresponding to the same sky direction, the transported polarization directions are expected to be identical.

The weighting term is defined as $w_{tj}=\sqrt{d_{\mathrm{nom},tj}d_{\mathrm{nom},0j}}$, where $d_{\mathrm{nom}}$ denotes the pre-calibration DoLP estimate. 
This weighting reduces the influence of observations with weak linear polarization, where the estimated AoLP is unstable.

\subsection{Complementary Residuals and Joint Optimization}
\label{sec:optimization}
The calibration parameters are jointly estimated by minimizing three complementary terms: 
cross-frame polarization consistency, four-channel measurement consistency, and a lens-response regularization term. 
The first term enforces polarization invariance across observations, while the latter two constrain the analyzer and lens models.

\textbf{Four-channel consistency.}
For a rank-three analyzer matrix $\mathbf F$, the four measured channels provide a one-dimensional null-space constraint. The corresponding reconstruction residual is defined as
\begin{equation}
r^c_o=\mathbf n^{T}\mathbf i_o,
\qquad
\mathbf n^{T}\mathbf F=\mathbf 0,
\qquad
\|\mathbf n\|_2=1.
\label{eq:channel_residual}
\end{equation}
where $\mathbf n$ is the unit left-null vector of $\mathbf F$ and is updated according to the current analyzer parameters. 
With equally weighted channels, this residual is equivalent to $(r^c_o)^2=\|\mathbf i_o-\mathbf F\mathbf D\mathbf i_o\|_2^2$.
This term constrains the analyzer response independently of the cross-frame polarization residual.

\textbf{Lens-response penalty.}
To prevent excessive attenuation and unstable inversion of the lens response, we impose a soft constraint on $B(\theta)$. 
For $L$ uniformly sampled incidence angles within the observed field of view, the penalty is defined as
\begin{equation}
r^B_\ell
=\kappa\min\{B(\theta_\ell)-B_m,0\},
\qquad \ell=1,\ldots,L,
\label{eq:lens_penalty}
\end{equation}
where $B_m>0$ is the minimum allowed response margin and $\kappa$ controls the penalty strength.

\textbf{Joint optimization.}
Let $\mathcal A$ denote the set of reference--current frame correspondences and $\mathcal O$ the set of observations. 
The calibration parameters are obtained by solving
\begin{equation}
\begin{aligned}
\boldsymbol\eta^*=\arg\min_{\boldsymbol\eta\in\Omega}\bigg\{
&\sum_{(t,j)\in\mathcal A}\sum_{h=1}^{2}\rho_f(r^a_{tj,h})\\
&+\sum_{o\in\mathcal O}\rho_f(\sqrt{\lambda_c}\,r^c_o)
+\sum_{\ell=1}^{L}\rho_f(r^B_\ell)\bigg\},
\end{aligned}
\label{eq:objective}
\end{equation}
where $\Omega$ defines the feasible parameter range and $r^a_{tj,h}$ denotes the $h$-th component of the cross-frame polarization residual.

We employ the soft-$\ell_1$ loss
\begin{equation}
\rho_f(r)=f^2\left(\sqrt{1+(r/f)^2}-1\right),
\label{eq:robust_loss}
\end{equation}
where $f>0$ controls the transition from quadratic to approximately linear penalization. 
This robust formulation reduces the influence of erroneous correspondences and measurement outliers during calibration.

\section{Experiment}
\label{sec:experiments}

\subsection{Experimental Setup}

\begin{figure}[!htbp]
\centering
\includegraphics[width=\columnwidth]{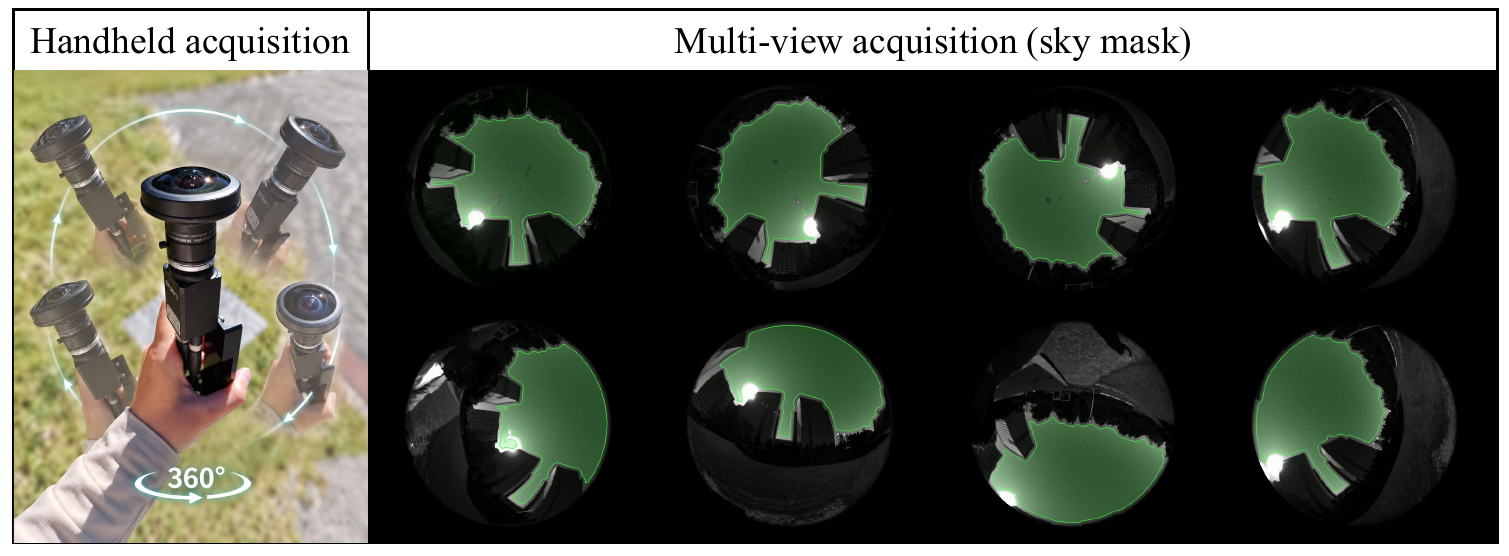}
\caption{Handheld outdoor data acquisition and representative sky-masked images captured from different viewing orientations.}
\label{fig:handheld_acquisition}
\end{figure}

\textbf{Hardware and data.}
In our method, only a short handheld rotation sequence under clear-sky conditions is needed to collect partially overlapping sky observations from multiple viewing orientations, without dedicated laboratory calibration equipment. Fig.~\ref{fig:handheld_acquisition} illustrates the acquisition procedure and representative views.

We use a LUCID TRI050s-QC color polarization camera with a Fujinon FE185C057HA fisheye lens. Two independent sequences are collected: \textit{Data A} (142 images in 9.5s) for parameter estimation and \textit{Data B} (159 images in 10.6s) for evaluation with fixed calibration parameters. Images are recorded as 8-bit TIFF mosaics at $2448\times2048$ pixels, yielding a $612\times512$ four-analyzer grid per color channel. The first image in each sequence serves as the reference frame.

\textbf{Implementation and evaluation protocol.}
We retain valid polarization measurements with $S_0\geq4$ DN, and $0.3<\mathrm{DoLP}\leq1$. The proposed model estimates eight parameters, where $g_0=1$ and $\delta_0=0$ fix the reference channel. All parameters are constrained within physically valid ranges, as summarized in Tab.~\ref{tab:parameter_bounds}. 

\begin{table}[!htbp]
\caption{Parameter bounds used in optimization.}
\label{tab:parameter_bounds}
\centering
\renewcommand{\arraystretch}{1.2}
\setlength{\tabcolsep}{8pt}
\begin{tabular*}{0.82\columnwidth}{@{\extracolsep{\fill}}lc}
\hline
\textbf{Parameter} & \textbf{Range} \\
\hline
$k_2$ & $[-1,1]$\\
$k_4$ & $[-0.5,0.5]$\\
Analyzer offsets & $[-5^\circ,5^\circ]$\\
Channel gains & $[0.9,1.1]$\\
\hline
\end{tabular*}
\end{table}

\begin{figure*}[!b]
\centering
\includegraphics[width=\textwidth]{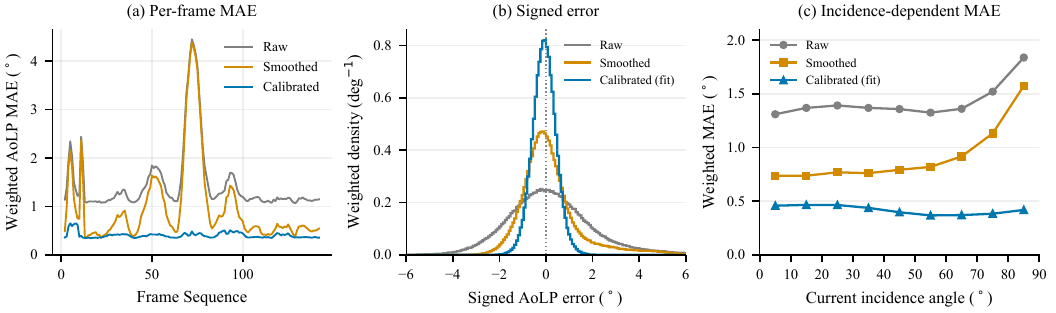}
\caption{Blue-channel calibration results on \textit{Data A}. (a) Weighted AoLP MAE. (b) Weighted signed-error distributions. (c) Weighted AoLP MAE in $10^\circ$ incidence-angle bins. All stages use identical observation pairs and fixed nominal DoLP weights.}
\label{fig:experiment_consistency}
\end{figure*}

We evaluate cross-frame consistency using the transported AoLP difference between corresponding sky observations. The angular residuals are wrapped to $[-90^\circ,90^\circ)$, and the weighted MAE and RMSE are computed as
\begin{equation}
\mathrm{MAE}=\frac{\sum_i w_i|e_i|}{\sum_i w_i},\qquad
\mathrm{RMSE}=\sqrt{\frac{\sum_i w_i e_i^2}{\sum_i w_i}},
\label{eq:experiment_metrics}
\end{equation}
where $e_i$ denotes the signed AoLP residual and $w_i$ is the nominal DoLP-based weight computed from the smoothed observations before calibration.

\subsection{Calibration and Cross-Sequence Validation}
We first estimate the blue-channel lens--analyzer parameters and evaluate calibration performance on \textit{Data A}, then assess their transfer to \textit{Data B} without refitting.

\textbf{Calibration on \textit{Data A}.}
The eight-parameter model fitted on \textit{Data A} converges without activating the parameter bounds:
\begin{equation}
\begin{aligned}
\boldsymbol{\eta}
={}&[-0.017876,\ 0.000154,\ -0.098^\circ,\ -0.042^\circ,\\
&\phantom{[}-0.086^\circ,\ 1.0180,\ 0.9418,\ 1.0081].
\end{aligned}
\label{eq:fitted_parameters}
\end{equation}

Fig.~\ref{fig:blue_fitted_response}(a) illustrates the incidence geometry of the fisheye polarization camera, with $\theta$ measured relative to the optical axis. Fig.~\ref{fig:blue_fitted_response}(b) shows the fitted blue-channel Mueller matrix parameter for the local incidence plane, with $B(90^\circ)=0.9568$. At this incidence angle, fully linearly polarized light oriented at $45^\circ$ to the local $p$ axis undergoes an AoLP shift to $43.74^\circ$, corresponding to a deviation of approximately $1.26^\circ$. The fitted analyzer gains deviate from unity by at most 5.82\%, while all angle offsets remain below $0.10^\circ$ in magnitude. Together, these parameters characterize the effective lens--analyzer response.

Fig.~\ref{fig:experiment_consistency} summarizes calibration performance on \textit{Data A}. Smoothing reduces the weighted cross-frame AoLP MAE from $1.406^\circ$ to $0.896^\circ$, and calibration further reduces it to $0.404^\circ$, a 54.9\% improvement over the smoothed baseline. The per-frame results in Fig.~\ref{fig:experiment_consistency}(a) show lower calibrated MAE for all 141 evaluated frames, while the signed-error distributions in Fig.~\ref{fig:experiment_consistency}(b) become more concentrated around zero. Fig.~\ref{fig:experiment_consistency}(c) examines the dependence of MAE on incidence angle, showing that calibration substantially suppresses the increase in error at larger angles. Specifically, MAE decreases from $0.755^\circ$ to $0.464^\circ$ in the central region ($0^\circ\leq\theta<30^\circ$) and from $1.573^\circ$ to $0.419^\circ$ in the peripheral region ($80^\circ\leq\theta<90^\circ$), corresponding to reductions of 38.5\% and 73.4\% relative to the smoothed baseline, respectively. These results demonstrate improved cross-frame consistency in both the central and peripheral image regions.

\begin{figure*}[!t]
\centering
\includegraphics[width=\textwidth]{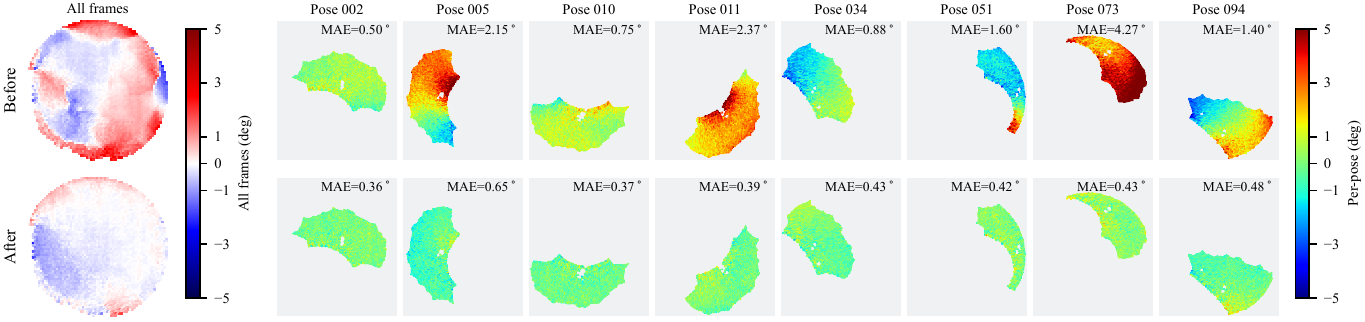}
\caption{Spatial cross-frame AoLP errors on \textit{Data A} before and after calibration. The upper and lower rows show smoothed, uncalibrated and calibrated results, respectively. The first column shows weighted mean signed errors, while the remaining columns show pairwise signed errors for eight representative frames relative to the reference frame.}
\label{fig:experiment_map}
\end{figure*}

Fig.~\ref{fig:experiment_map} further visualizes the spatial distribution of signed cross-frame AoLP errors on \textit{Data A}. The first column aggregates measurements across all frames, showing substantially reduced spatial biases after calibration. Higher residual errors in the lower-left region mainly reflect sparse observations, as high-DoLP measurements were concentrated on the right side during acquisition. The remaining columns present eight representative frames, where the calibrated results exhibit lower weighted MAE and reduced signed errors over the evaluated sky regions. These results further demonstrate improved spatial cross-frame AoLP consistency on the calibration sequence.

\begin{figure}[!htbp]
\centering
\includegraphics[width=0.76\columnwidth]{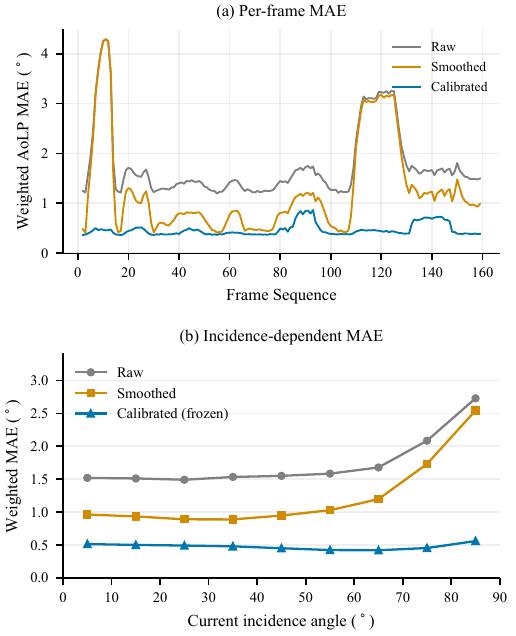}
\caption{Blue-channel validation on \textit{Data B} using parameters estimated from \textit{Data A} without refitting. (a) Weighted AoLP MAE. (b) Weighted AoLP MAE in $10^\circ$ incidence-angle bins.}
\label{fig:blue_dataB_validation}
\end{figure}

\textbf{Validation on \textit{Data B}.}
To evaluate cross-sequence transfer, we apply the parameters estimated from \textit{Data A} to \textit{Data B}, acquired at a different time under cloudy skies, without refitting. Fig.~\ref{fig:blue_dataB_validation}(a) shows that calibration improves all 158 evaluated frames relative to the smoothed baseline. Overall, smoothing reduces the weighted cross-frame AoLP MAE from $1.710^\circ$ to $1.189^\circ$, and calibration further reduces it to $0.466^\circ$, a 60.8\% improvement over the smoothed baseline. The localized residual peaks near sequence positions 90 and 140 are associated with reduced feature support at viewpoints far from the reference. Fig.~\ref{fig:blue_dataB_validation}(b) examines the dependence of MAE on incidence angle, showing that calibration substantially reduces the large-angle errors. Specifically, MAE decreases from $0.914^\circ$ to $0.498^\circ$ in the central region ($0^\circ\leq\theta<30^\circ$) and from $2.542^\circ$ to $0.560^\circ$ in the peripheral region ($80^\circ\leq\theta<90^\circ$), corresponding to reductions of 45.5\% and 78.0\% relative to the smoothed baseline, respectively. Although residual errors still increase near the field boundary, the consistent improvements across frames and image regions support the transferability of the estimated response parameters across the tested acquisition conditions.

\begin{figure}[!htbp]
\centering
\includegraphics[width=0.85\columnwidth]{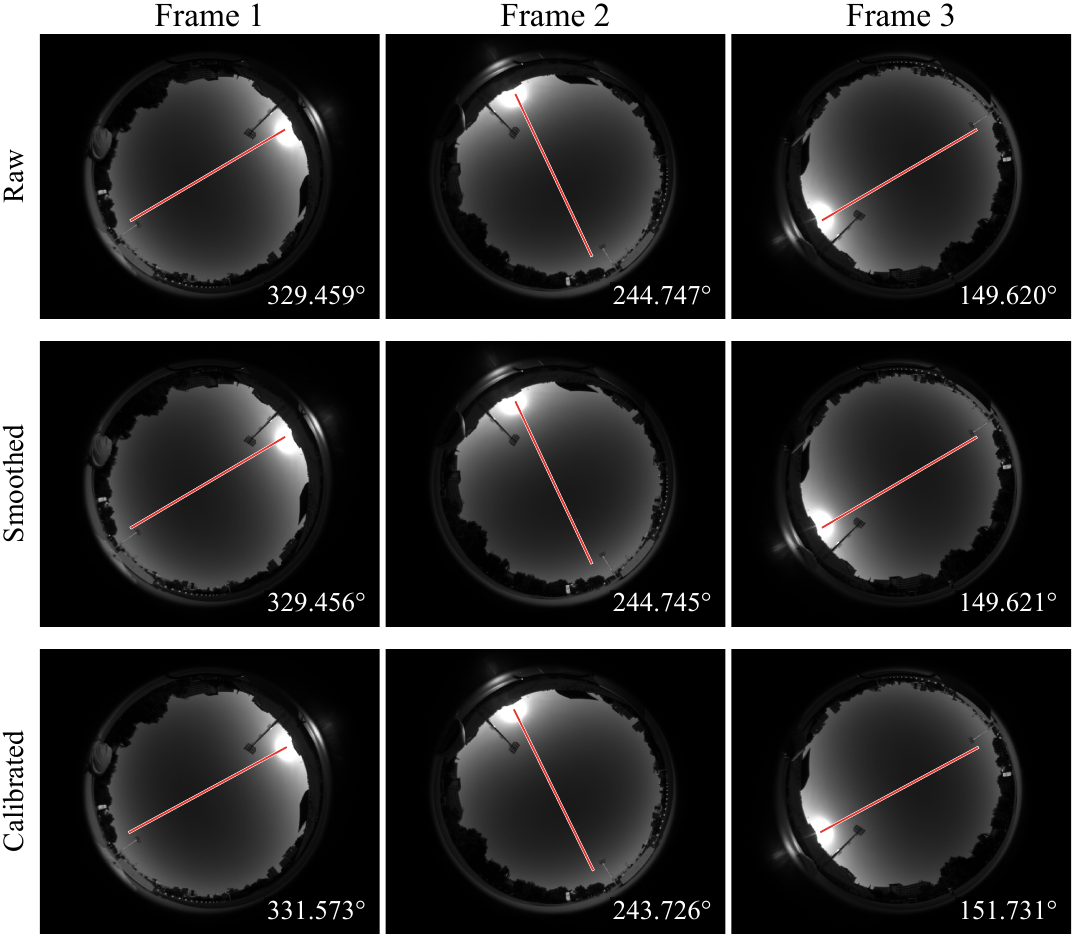}
\caption{Preliminary solar-azimuth validation. Red lines indicate estimated solar azimuths in the camera frame.}
\label{fig:solar_azimuth_validation}
\end{figure}

\textbf{Solar-Azimuth Validation.}
\label{sec:solar_validation}
To preliminarily assess the utility of the proposed calibration for polarization navigation, we evaluate solar-azimuth estimation using outdoor observations collected with the polarization camera mounted on a Unitree Go2 quadruped robot. Image-based navigation can recover a directional reference from pixel-wise skylight AoLP measurements~\cite{courtier2022image}. In our evaluation, the AoLP at each valid sky pixel is combined with its viewing geometry to construct a polarization direction vector, and the solar direction is estimated by enforcing orthogonality to these vectors. Its azimuth is then computed in the camera frame. The same estimator and valid pixel set are used for raw, smoothed, and calibrated measurements, allowing the effect of AoLP correction on solar-azimuth recovery to be examined.

Fig.~\ref{fig:solar_azimuth_validation} compares the estimated solar-azimuth lines for three representative frames. After calibration, the estimated lines align more closely with the visible solar center, whereas smoothing alone produces little change. Because the estimator operates on pixel-wise AoLP values, this improvement provides preliminary evidence that correcting the lens--analyzer response benefits a downstream navigation task. More accurate solar-azimuth recovery can provide a more reliable reference for relative-heading estimation, supporting polarization navigation and providing a basis for future integration into a polarization--inertial navigation system.

\subsection{Ablation Study}

\begin{table*}[!t]
\caption{Component ablation on the Blue channel data. S: intensity smoothing; L: lens response; D: analyzer-angle offsets; G: relative gains; 'w/o': without. }
\label{tab:experiment_ablation}
\centering
\begingroup
\small
\renewcommand{\arraystretch}{1.15}
\setlength{\tabcolsep}{5pt}
\begin{tabular*}{\textwidth}{@{\extracolsep{\fill}}lcccccrrrr@{}}
\hline
 & \multicolumn{4}{c}{Components}
 & & \multicolumn{2}{c}{Data A: fitting}
 & \multicolumn{2}{c}{Data B: validation} \\
\cline{2-5}\cline{7-8}\cline{9-10}
Variant & S & L & D & G & Params.
& MAE & RMSE & MAE & RMSE \\
\hline
Raw
& -- & -- & -- & -- & 0
& 1.406 & 1.858 & 1.710 & 2.209 \\
S
& $\checkmark$ & -- & -- & -- & 0
& 0.896 & 1.330 & 1.189 & 1.657 \\
S+L
& $\checkmark$ & $\checkmark$ & -- & -- & 2
& 0.667 & 0.913 & 0.882 & 1.173 \\
S+L+D
& $\checkmark$ & $\checkmark$ & $\checkmark$ & -- & 5
& 0.574 & 0.789 & 0.686 & 0.907 \\
\textbf{Full}
& $\checkmark$ & $\checkmark$ & $\checkmark$ & $\checkmark$ & 8
& \textbf{0.404} & \textbf{0.515}
& \textbf{0.466} & \textbf{0.616} \\
\hline
Full w/o S
& -- & $\checkmark$ & $\checkmark$ & $\checkmark$ & 8
& 1.098 & 1.402 & 1.176 & 1.500 \\
Full w/o L
& $\checkmark$ & -- & $\checkmark$ & $\checkmark$ & 6
& 0.596 & 0.788 & 1.040 & 1.437 \\
Full w/o D
& $\checkmark$ & $\checkmark$ & -- & $\checkmark$ & 5
& 0.404 & 0.515 & 0.470 & 0.621 \\
\hline
\end{tabular*}
\endgroup
\end{table*}

We conduct an ablation study to assess the contribution of each component to calibration accuracy and transfer performance. Tab.~\ref{tab:experiment_ablation} compares intensity smoothing (S), lens response (L), analyzer-angle offsets (D), and relative gains (G). Each model is fitted independently on \textit{Data A} and tested on \textit{Data B} using fixed parameters. After smoothing, successively adding L, D, and G reduces the test MAE from $1.189^\circ$ to $0.882^\circ$, $0.686^\circ$, and $0.466^\circ$, with corresponding improvements on \textit{Data A}. These improvements are consistent with the physical formulation of the calibration model. Smoothing suppresses intensity noise, while the lens and analyzer parameters capture incidence-dependent polarization distortion, angular misalignment, and sensitivity imbalance, jointly improving fitting and test accuracy.

The last three rows of Tab.~\ref{tab:experiment_ablation} remove one component at a time from the full model, with 'w/o' denoting 'without', and refit the remaining parameters. These comparisons highlight the role of the lens model in transfer across acquisition conditions. Without L, MAE increases from $0.404^\circ$ to $0.596^\circ$ on \textit{Data A} and from $0.466^\circ$ to $1.040^\circ$ on \textit{Data B}, widening the fitting-to-test gap from $0.062^\circ$ to $0.444^\circ$. Removing S produces high errors on both datasets ($1.098^\circ$ and $1.176^\circ$), while removing D has little effect on the test MAE. These results indicate that smoothing is important for measurement precision and that the lens model is particularly important for maintaining accuracy across the tested acquisition conditions.

\subsection{Discussion}
\label{sec:discussion}
We further examine the robustness and effectiveness of the proposed calibration method by evaluating its performance with different numbers of calibration frames and analyzing the spatial distribution of residual errors.

\begin{table}[!htbp]
\caption{Response fitting with subsets of Data A.}
\label{tab:observation_count}
\centering
\begingroup
\small
\renewcommand{\arraystretch}{1.15}
\setlength{\tabcolsep}{4pt}
\begin{tabular*}{\columnwidth}{@{\extracolsep{\fill}}rrrr@{}}
\hline
Total frames & Fitting pairs & A MAE & B MAE \\
\hline
5 & 13,559 & 0.409 & 0.474 \\
9 & 31,256 & 0.413 & 0.552 \\
17 & 59,587 & 0.406 & 0.470 \\
33 & 117,668 & 0.405 & 0.472 \\
142 (all) & 517,983 & 0.404 & 0.466 \\
\hline
\end{tabular*}
\endgroup
\end{table}
\label{sec:observation_count}

\textbf{Minimum Frame Requirements for Fitting.}
We refit the eight-parameter response using the reference frame and 4, 8, 16, or 32 uniformly sampled current frames from \textit{Data A}. As shown in Tab.~\ref{tab:observation_count}, five frames, the smallest subset tested, achieve MAEs of $0.409^\circ$ and $0.474^\circ$ on all \textit{Data A} and \textit{Data B} pairs, close to the full-sequence results of $0.404^\circ$ and $0.466^\circ$. This subset retains optical-axis tilts of $54.6^\circ$ and $65.2^\circ$, providing diverse incidence angles and thousands of pixel correspondences. The nine-frame subset contains mostly small-tilt views, and its weaker large-angle coverage is consistent with its higher transfer error. These results suggest that representative pose coverage enables accurate response fitting with few frames.

\begin{figure}[!htbp]
\centering
\includegraphics[width=0.90\columnwidth]{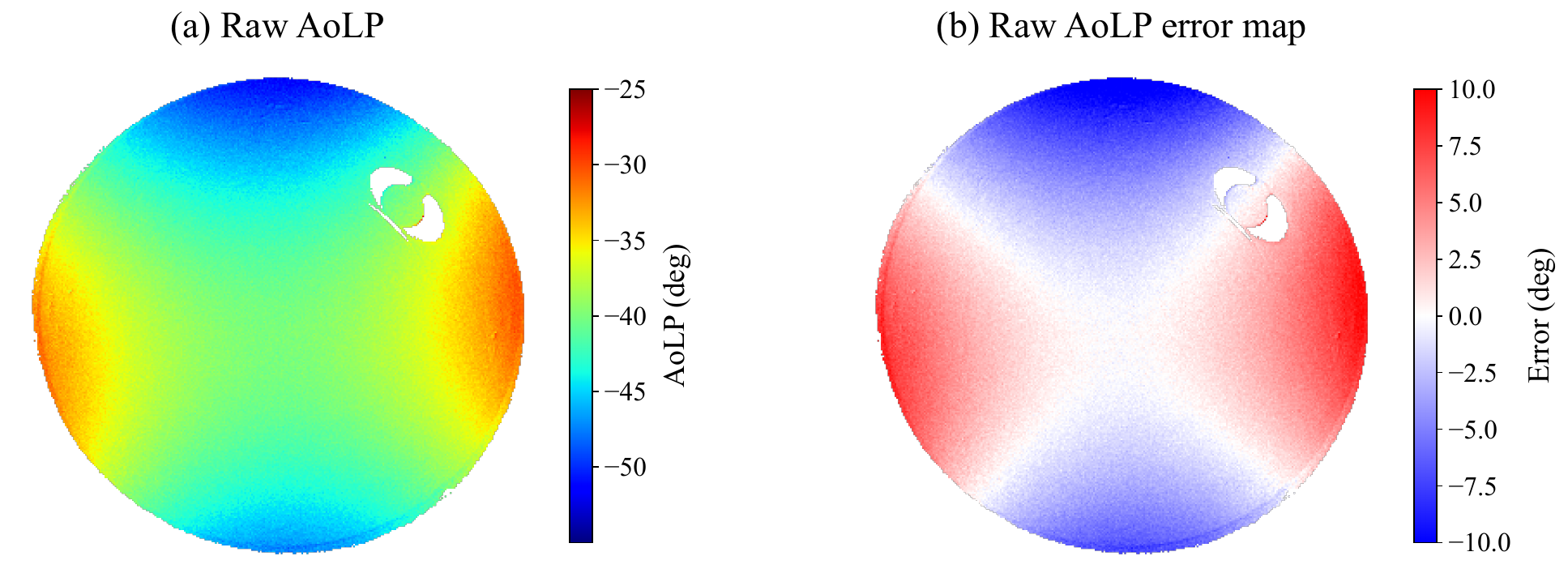}
\caption{Integrating-sphere measurements over a partial fisheye field of view: (a) raw AoLP and (b) raw AoLP error distribution.}
\label{fig:integrating_sphere_error}
\end{figure}

\begin{figure*}[!t]
\centering
\includegraphics[width=\textwidth]{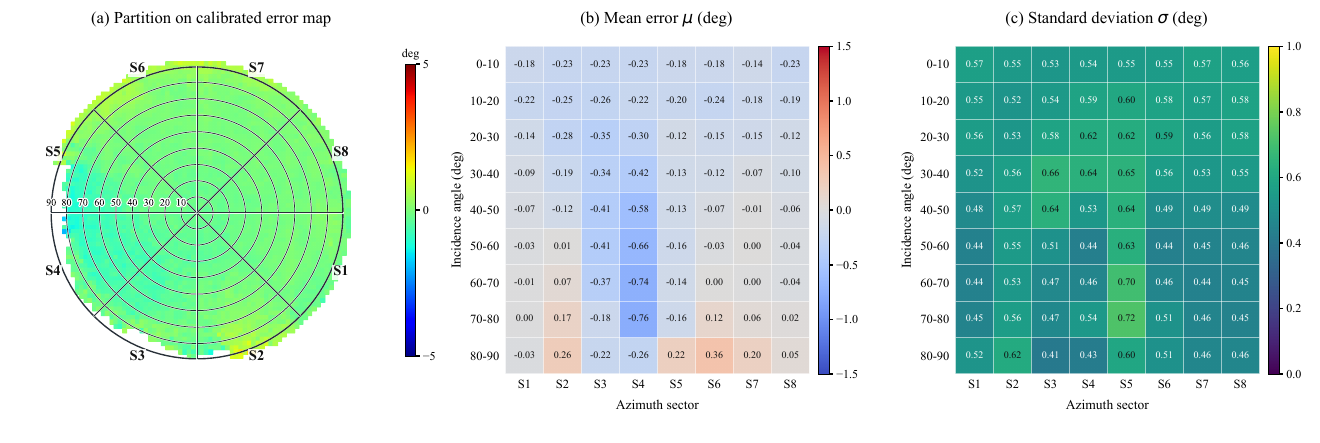}
\caption{Spatial analysis of calibrated AoLP residuals on \textit{Data A}. (a) Nine $10^\circ$ incidence rings and eight $45^\circ$ azimuth sectors overlaid on the all-frame error map. (b) Mean signed residual and (c) standard deviation in each block.}
\label{fig:calibration_error_analysis}
\end{figure*}

\textbf{Calibration Error Analysis.}
\label{sec:calibration_error_analysis}
Fig.~\ref{fig:integrating_sphere_error} shows the AoLP error distribution measured under integrating-sphere illumination over only part of the fisheye field of view, exhibiting an approximately symmetric pattern consistent with the lens's intrinsic response, while also containing errors introduced by oblique incidence and other experimental setup factors.

To examine the remaining spatial errors after calibration, we pool the calibrated cross-frame AoLP residuals over 72 spatial blocks of \textit{Data A}, as shown in  Fig.~\ref{fig:calibration_error_analysis}(a). The blockwise mean residuals and standard deviations in Fig.~\ref{fig:calibration_error_analysis}(b) and (c) reveal small spatial biases and broadly consistent residual spread, with standard deviations of $0.41^\circ$–$0.72^\circ$ and a median of $0.54^\circ$, suggesting that the spatially structured distortions associated with the lens response have been largely corrected. Remaining local biases may partly reflect uneven observation coverage, with high-DoLP measurements concentrated on the right side and fewer constraints available in the lower-left region. Cloud cover and partial sky occlusion can also introduce scene-dependent errors beyond camera-response calibration, potentially contributing to the higher validation error on \textit{Data B}. More balanced spatial coverage and diverse viewing orientations could further strengthen calibration constraints in undersampled regions.

\section{Conclusion}
\label{sec:conclusion}
We presented a sky-based calibration method for jointly correcting fisheye lens and analyzer responses without dedicated polarization references. The proposed method effectively calibrates polarization measurements across both central and peripheral regions of the fisheye field of view. Preliminary solar-azimuth validation demonstrates its potential for polarization-based navigation. Future work will focus on refining the response model and integrating the calibration into navigation systems.

\balance
\bibliographystyle{IEEEtran}
\bibliography{references}

\end{document}